\documentclass[letterpaper]{article} % DO NOT CHANGE THIS
\usepackage[preprint]{aaai2027}  % DO NOT CHANGE THIS
\usepackage[hyphens]{url}  % DO NOT CHANGE THIS
\usepackage{graphicx} % DO NOT CHANGE THIS
\usepackage{natbib}  % DO NOT CHANGE THIS AND DO NOT ADD ANY OPTIONS TO IT
\usepackage{caption} % DO NOT CHANGE THIS AND DO NOT ADD ANY OPTIONS TO IT
\usepackage{algorithm}
\usepackage{algorithmic}
\usepackage{multirow}
\usepackage{tabularx}
\usepackage{array}
\usepackage{hyperref}
\newcolumntype{C}{>{\centering\arraybackslash}X}
\usepackage{newfloat}
\usepackage{listings}
\DeclareCaptionStyle{ruled}{labelfont=normalfont,labelsep=colon,strut=off} % DO NOT CHANGE THIS
\floatstyle{ruled}
\newfloat{listing}{tb}{lst}{}
\floatname{listing}{Listing}

\usepackage{booktabs}
\usepackage{amsmath} 
\usepackage{amssymb} 
\title{Domain Generalization for Style-Debiased Object Detection via Label-Consistent Training across Multiple Styles}
\author{
    Chaoran Zhang\textsuperscript{\rm 1},
    Shirshak Bhusal\textsuperscript{\rm 1},
    Zhilin Gao\textsuperscript{\rm 1},
    Qiang Xue\textsuperscript{\rm 2},
    Haoting Shen\textsuperscript{\rm 1}\corresponding
}
\affiliations{
    \textsuperscript{\rm 1}College of Computer Science and Technology, Zhejiang University\\
    \textsuperscript{\rm 2}Data security,\\
    Hangzhou High-Tech Zone (Binjiang) Institute of Blockchain and Data Security
}

\begin{document}

\maketitle

\begin{abstract}
% Domain generalization (DG) for object detection aims to improve model performance on unseen target domains. Existing detectors, however, suffer substantial degradation on visually distinct unseen domains, as models confined to limited training styles tend to overfit superficial statistical cues rather than object-level semantics. Motivated by human perception, which attends primarily to semantically relevant features while ignoring irrelevant style cues, we propose a cross-domain consistency learning framework that couples cross-domain data synthesis with consistency-driven training to suppress style-sensitive shortcut learning. For data preparation, we leverage off-the-shelf diffusion models to expand the source datasets into 20 distinct styles, with all generated images manually verified to ensure object instances remain consistent across style variants. During training, we further incorporate consistency constraints into the loss design to reinforce the learning of cross-domain stable features for each scene while suppressing the influence of style-sensitive cues. Extensive experiments and ablation studies on multiple challenging DG benchmarks show that our method achieves state-of-the-art performance, even surpassing numerous Domain adaption approaches. Notably, our approach applies to both single-stage and two-stage detectors and is broadly compatible with other DG methods. The training code and two style-transfer datasets constructed for the experiments, comprising approximately 180K images in total, are publicly available on GitHub.

Domain generalization (DG) for object detection aims to improve model
performance on unseen target domains. Existing detectors, however, suffer
substantial degradation on visually distinct unseen domains, as models
confined to limited training styles tend to overfit superficial statistical
cues rather than object-level semantics. Motivated by human perception,
which attends primarily to semantically relevant features while ignoring
irrelevant style cues, we propose Origin-Anchored Multi-Style Consistency
(OAMSC), a source-only framework that couples multi-style data synthesis with
cross-style consistency learning to suppress style-sensitive shortcut
learning. For data preparation, we leverage off-the-shelf diffusion models
to expand the source datasets into 20 distinct styles, with all generated
images manually verified to ensure that object instances and annotations
remain consistent across style variants. During training, we impose
Origin-anchored consistency constraints at scene and object levels to
reinforce the learning of domain-stable features while suppressing the
influence of style-sensitive cues. Extensive experiments and ablation studies on multiple challenging DG
benchmarks demonstrate the consistent effectiveness and strong
generalization performance of OAMSC across diverse domain shifts. Notably, our approach applies to both one-stage and two-stage
detectors, while its data-centric design is complementary in principle to
other DG methods. The training code and two style-transfer datasets
constructed for the experiments, comprising approximately 180K images in
total, are publicly available \href{https://anonymous.4open.science/r/OAMSC-C07A/}{on GitHub}.
\end{abstract}


\section{Introduction}
\label{sec:introduction}

Object detection models have achieved near-saturating performance on curated benchmarks, yet their accuracy degrades drastically in real-world deployment due to ubiquitous domain shifts caused by sensor variations, illumination changes, style discrepancies, and background differences \cite{zhou2023dg}. In high-stakes scenarios such as industrial inspection, autonomous driving, and remote sensing, annotating sufficient target-domain data is often prohibitively costly or infeasible. Domain generalization (DG) for object detection, which aims to learn robust detectors solely from source-domain data that generalize to unseen target domains without target data access, has thus emerged as a critical research direction to bridge the gap between laboratory results and practical deployment \cite{zhou2023dg}.

Existing DG detection methods can be broadly grouped into four complementary lines: data augmentation and domain diversification \cite{danish2024divalign, zhou2021mixstyle, lee2024oadg,qi2024doubleaug, rao2024srcd}, invariant representation learning and causal disentanglement \cite{xu2023mad, liu2024ufr}, meta-learning based optimization \cite{feng2024metalearning}, and foundation model-guided feature transfer \cite{vidit2023clip, he2025gdd}. Among them, the pipeline of data augmentation combined with consistency regularization stands out for industrial adoption, owing to its simplicity, detector-agnostic property, and seamless compatibility with standard detection systems.

Nevertheless, two limitations remain. Hand-crafted augmentations follow
predefined priors and may insufficiently cover real-world appearance shifts
\cite{danish2024divalign,rao2024srcd}. Conventional consistency
regularization often lacks reliable scene- and object-level correspondence,
obscuring semantic invariance amid style and background changes. Unstable
references may also propagate augmentation artifacts
\cite{danish2024divalign}.

We therefore propose \emph{Origin-Anchored Multi-Style Consistency} (OAMSC),
a source-only DG framework using label-preserving multi-style data.
Off-the-shelf generative models render diverse appearances, while manual
verification preserves objects, geometry, and annotations. Synchronized
scene groups pair each stylized view with its \textsc{Origin}, and an EMA
\textsc{Origin} teacher anchors global FPN and GT-guided RoI consistency.
OAMSC promotes appearance-robust representations without changing detector
architecture or inference. Experiments across diverse shifts demonstrate
strong generalization.

Our main contributions are summarized as follows.
\begin{itemize}
    \item We introduce a same-scene multi-style consistency learning paradigm
for DG detection. Label-preserving variants serve as correlated views whose
correspondence supplies consistency supervision, preserving object semantics
while suppressing style-sensitive cues. This formulation exploits known
scene correspondence rather than treating stylized samples as unrelated
augmented images.

    \item Our scene-grouped training combines synchronized geometry with an
    EMA \textsc{Origin} teacher. Global FPN and GT-guided RoI alignment
    provide stable consistency at scene and object levels.

    \item Experiments on camera, synthetic-weather, simulation-to-real, and
    corruption shifts demonstrate compatibility with one- and two-stage
    detectors. We release the training code and two curated multi-style
    datasets containing approximately 180K generated images.
\end{itemize}

\section{Related Work}
\label{sec:related_work}
Domain generalization for object detection aims to learn robust detectors from labeled source domain(s) that generalize well to unseen target domains without accessing target data during training. Recent approaches can be broadly categorized into technical streams as follows. 

\textbf{Data augmentation and domain diversification} that enrich the input space with simulated perturbations. The strategy is to diversify the source domain through perturbations or randomizations, introducing designed variations to encourage the model to focus on invariant semantic structures rather than superficial statistics. For example, it was demonstrated \cite{danish2024divalign} that a carefully selected set of hand-crafted corruptions establishes a strong baseline for single-domain generalized detection. By enforcing the prediction consistency between original and augmented views via KL divergence on classification and L2 loss on localization, the detector-agnostic approach yielded noticeable improvements in out-of-domain accuracy and model calibration. Other works explored more specialized augmentation strategies: \cite{lee2024oadg} proposed object-aware mixing to generate multi-domain data while preserving semantic features of objects; \cite{qi2024doubleaug} introduced color perturbation and dual-style memory for urban-scene generalization; \cite{rao2024srcd} used texture-based self-augmentation coupled with semantic reasoning to maintain intrinsic semantic structures. Despite their simplicity and effectiveness, these methods rely on fixed, manually predefined perturbation sets, which inherently limits coverage of the full spectrum of domain shifts. Moreover, enforcing consistency via KL divergence can introduce over-regularization when augmentations become overly aggressive, often leading to a trade-off where source domain performance degrades in exchange for improved generalization.

\textbf{Feature extraction and causal inference.}
This direction explicitly disentangles causal (domain-invariant) features from non-causal (domain-specific) factors via architectural designs and causal theoretical guidance. Representative works include: \cite{liu2024ufr}, which built a structural causal model to analyze biases in single-source DG; \cite{xu2023mad}, which purified invariant features via multi-view adversarial discriminators; \cite{li2025cauvis}, which equated visual prompts with backdoor adjustment via cross-attention; and \cite{hong2026bridge}, which realized front-door adjustment via low-rank bases for vision foundation models. Despite theoretical rigor, these methods come with architectural complexity and computational overhead, rely on strong assumptions about causal graph structures, and inherit generalization limits when dependent on external tools like SAM.

\textbf{External information guidance.} 
This paradigm imports robust semantic priors from pre-trained foundation models. For examples, diffusion-based methods \cite{he2025gdd, he2025boosting} distilled intermediate diffusion features into lightweight detectors via multi-level alignment; CLIP-based works such as \cite{vidit2023clip} simulated unseen domains via text-driven semantic augmentation. However, diffusion methods prioritize geometric invariance and underperform when texture/color cues are meaningful, while CLIP-based approaches struggle with fine-grained object representation and incur high region-level alignment costs. Both rely heavily on pre-training quality and introduce notable architectural complexity.

\textbf{Statistical alignment and probabilistic modeling.}
These methods model domain relationships via distribution matching, targeting universal adaptation scenarios with unknown category overlap. Pioneered by \cite{shi2022usdaf} for universal adaptive detection, follow-ups include \cite{chen2022probabilistic} with uncertainty-guided consistency training, \cite{shi2024wadaptive} for weighted alignment and private class discovery, and \cite{zheng2025dpa} with dual probabilistic alignment for shared/private categories. The challenges include heavy computational overhead from memory banks and statistical operations, possibly fragile assumptions of clear category separability, and laborious hyperparameter tuning for varying domain shift magnitudes.

It is worth noting that the technical routes mentioned above are not mutually exclusive. An integrated technical frame that takes advantage of more than one route might be necessary for best practice.

% Required packages: amsmath, amssymb, bm, graphicx, booktabs, tabularx, array
% Suggested bibliography keys:
% ren2016faster, lin2017fpn, he2017maskrcnn,
% tarvainen2017meanteacher, gao2025seedream3, chen2025seedream4,
% bytedance2025seedream45, bytedance2026seedream5
% Fixed 9-pt Roman table bodies; captions remain controlled by the venue style.
\providecommand{\papertablefont}{\rmfamily\fontsize{9pt}{10.8pt}\selectfont}
\section{Method}
\label{sec:method}

\subsection{Overview}
\label{sec:method_overview}

We study source-only domain-generalized detection, where training uses
labeled source images but no target-domain data. Let the source set be
\begin{equation}
    \mathcal{D}_{\mathrm{src}}
    =
    \left\{
        \left(
            x_i^{0},
            y_i
        \right)
    \right\}_{i=1}^{N},
    \quad
    y_i
    =
    \left\{
        \left(
            b_{ij},
            c_{ij}
        \right)
    \right\}_{j=1}^{M_i},
    \label{eq:source_dataset}
\end{equation}
where $x_i^{0}$ is the original (\textsc{Origin}) image,
$b_{ij}$ and $c_{ij}$ denote the bounding box and class of object $j$,
respectively, and $M_i$ is the number of annotated objects.
Our \emph{Origin-Anchored Multi-Style Consistency} (OAMSC) retains Faster
R-CNN \cite{ren2016faster} and augments its training with scene-grouped
multi-style supervision, an EMA \textsc{Origin} teacher, and hierarchical
feature consistency.

\begin{figure*}[t]
    \centering
    % Replace the fallback box by placing the final overview at the path below.
    \IfFileExists{Figures/OAMSC_Overview.pdf}{
        \includegraphics[
            width=0.95\textwidth,
            keepaspectratio
        ]{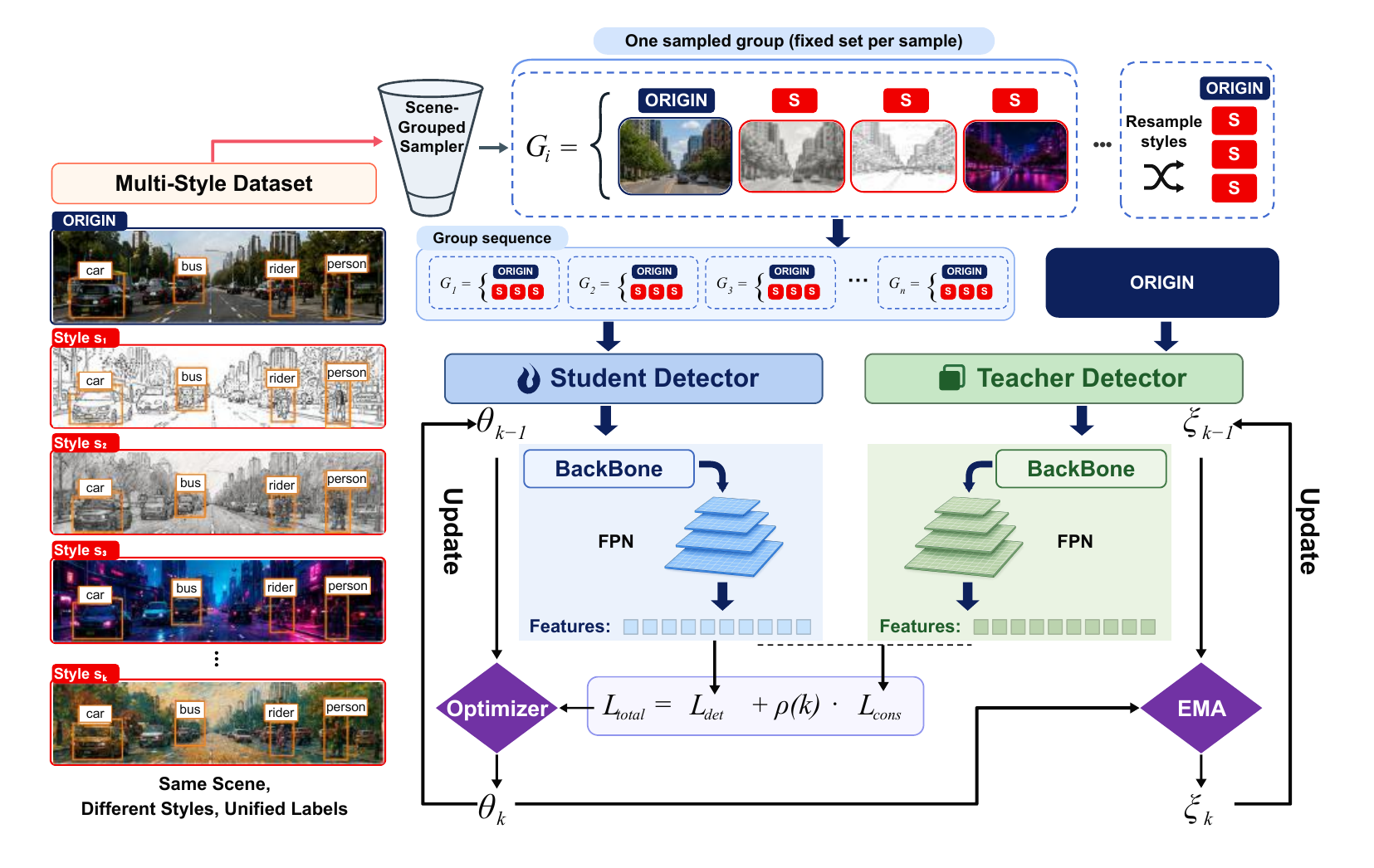}
    }{
        \fbox{
            \parbox[c][0.16\textheight][c]{0.96\textwidth}{
                \centering
                Placeholder for the OAMSC overview figure
            }
        }
    }
    \caption{Overview of OAMSC. Each group contains one \textsc{Origin} and
    three label-preserving styles. All views train the student, while
    \textsc{Origin} alone enters the EMA teacher as the consistency anchor.
    Back-propagation updates $\theta$ before EMA updates $\xi$;
    $\mathcal{L}_{\mathrm{cons}}$ combines global and GT-RoI consistency.}
    \label{fig:oamsc-overview}
\end{figure*}

We construct an offline bank of label-preserving views for each source scene.
The scene-grouped sampler forms
$\mathcal{G}_i=\{x_i^0,x_i^{s_1},x_i^{s_2},x_i^{s_3}\}$: the
\textsc{Origin} is fixed and the three styles are resampled across epochs.
One geometric transform is shared by the group and its boxes
(Eq.~\eqref{eq:synchronized_transform}); hence the teacher input is the same
synchronized \textsc{Origin}, not an independent view.

All group members receive detection supervision through the student, whereas
only \textsc{Origin} enters the teacher. Styled student features align with
detached teacher anchors at global FPN and GT-guided RoI levels. In
Figure~\ref{fig:oamsc-overview},
$\mathcal{L}_{\mathrm{cons}}\equiv
\lambda_{\mathrm{g}}\mathcal{L}_{\mathrm{global}}+
\lambda_{\mathrm{r}}\mathcal{L}_{\mathrm{roi}}$; its operations are defined in
Secs.~\ref{sec:global_alignment} and~\ref{sec:roi_alignment}.
At iteration $k$, back-propagation through
$\mathcal{L}_{\mathrm{det}}+\rho(k)\mathcal{L}_{\mathrm{cons}}$ updates only
the student from $\theta_{k-1}$ to $\theta_k$; EMA then updates the detached
teacher from $\xi_{k-1}$ to $\xi_k$ for the next iteration. This asymmetry
provides a slowly evolving photographic reference for diverse student views.

\subsection{Label-Preserving Multi-Style Data}
\label{sec:multistyle_construction}

Using the commercial Seedream 3.0, 4.0, 4.5, and 5.0 Lite models
\cite{seedream3,seedream4,seedream45,seedream5}, we render each source image
from Cityscapes~\cite{cordts2016cityscapes} and
RealDriveSim~\cite{jadon2025realdrivesim} in 20 styles. This produces
179{,}471 variants (approximately 180K), excluding \textsc{Origin}, with no
generation cost during detector training or inference.

\begin{figure*}[t]
    \centering
    \IfFileExists{Figures/Multi_style_Images_Sample.png}{
        \includegraphics[
            width=0.98\textwidth,
            keepaspectratio
        ]{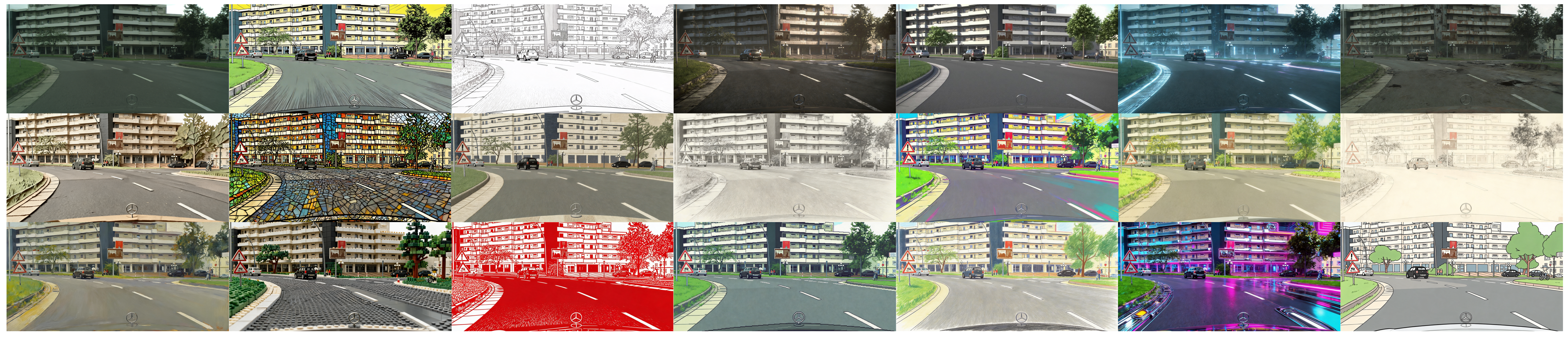}
    }{
        \fbox{
            \parbox[c][0.16\textheight][c]{0.96\textwidth}{
                \centering
                Placeholder for the 21-view style visualization
            }
        }
    }
    \caption{One 21-view example from Cityscapes: the
    \textsc{Origin} and 20 style variants preserve scene geometry, object
    identity and location, and therefore share the same annotations.}
    \label{fig:style-visualization}
\end{figure*}

We select five styles from each of four families: painting and fine art,
anime and comics, film and games, and materials and crafts
(Table~\ref{tab:style-selection}). Every variant is manually compared with
its \textsc{Origin}; missing, duplicated, displaced, or deformed objects and
geometry or annotation mismatches trigger regeneration and reinspection.
The accepted variants therefore reuse the original boxes and classes, as
illustrated in
Figure~\ref{fig:style-visualization}.

\begin{table*}[t]
    \centering
    \papertablefont
    \setlength{\tabcolsep}{1.5pt}
    \renewcommand{\arraystretch}{1.12}
    \begin{tabularx}{0.96\textwidth}{
        >{\raggedright\arraybackslash}p{0.2\textwidth}
        C C C C C
    }
        \toprule
    
        \multicolumn{1}{c}{\textbf{Category}}
        &
        \multicolumn{5}{c}{\textbf{Selected Styles}}
        \\

        \midrule
        % \cmidrule(lr){1-6}

        Painting and Fine Art &
        Oil Painting&
        Sketching        &
        Line Art        &
        Crayon Drawing        &
        Chinese Painting        \\

        Anime and Comics        &
        Ghibli Anime        &
        American Comics       &
        Chibi Comics        &
        Painterly Anime        &
        Pixel Art        \\

        Film and Games       &
        3D Modeling        &
        Post-Apocalyptic        &
        Science-Fiction        &
        Cyberpunk        &
        AAA Game Scene        \\

        Materials and Crafts        &
        Paper Cutting       &
        Stained Glass        &
        Building Blocks        &
        Collage        &
        Textile Art        \\

        \bottomrule
    \end{tabularx}

    \caption{The 20 appearance styles in the complete candidate bank,
    organized into four semantic families. \textsc{Origin} is retained
    separately and is not counted as a transformed style.}
    \label{tab:style-selection}
\end{table*}

Let $\mathcal{S}$ be a selected subset of the bank,
$K=|\mathcal{S}|\leq20$, and let
$\mathcal{S}_i\subseteq\mathcal{S}$ contain the available styles for scene
$i$. The transformed views, view pool, and inherited labels are
\begin{equation}
    \begin{aligned}
        x_i^s &= \mathcal A_s(x_i^0), \qquad s\in\mathcal S_i,\\
        \mathcal V_i
        &= \{x_i^0\}\cup\{x_i^s\mid s\in\mathcal S_i\},\\
        y_i^0 &= y_i^s = y_i, \qquad s\in\mathcal S_i,
    \end{aligned}
    \label{eq:style_bank}
\end{equation}
where $\mathcal A_s$ preserves labels and geometry.

Views are indexed by scene after removing unavailable pairs and invalid
boxes. Scene identity, annotation index, and class then provide explicit
cross-style correspondence. The bank serves as a sampling pool, and every
sampled view retains detection supervision.

\subsection{Origin-Anchored Grouped Training}
\label{sec:origin_anchor}

\subsubsection{Scene-Grouped View Sampling}
\label{sec:scene_group_sampling}

We sample scenes rather than independent images and form
\begin{equation}
    \mathcal{G}_i
    =
    \left\{
        x_i^{0},
        x_i^{s_1},
        \ldots,
        x_i^{s_m}
    \right\},
    \quad
    \left\{
        s_1,\ldots,s_m
    \right\}
    \subseteq
    \mathcal{S}_i,
    \label{eq:scene_group}
\end{equation}
where $m=3$. Styles are resampled each epoch and, when possible, drawn from
distinct families in Table~\ref{tab:style-selection}, preserving a
photographic anchor while fixing per-step cost.

\subsubsection{Synchronized Geometric Augmentation}
\label{sec:synchronized_geometry}

To preserve correspondence, one geometric transform $T_i$ is shared by all
views and annotations in $\mathcal{G}_i$:
\begin{equation}
    \widetilde{x}_i^{s}=T_i(x_i^{s}),
    \quad
    \widetilde{b}_{ij}=T_i(b_{ij}),
    \quad
    s\in\{0,s_1,\ldots,s_m\}.
    \label{eq:synchronized_transform}
\end{equation}
The shared horizontal flip ensures that consistency measures appearance
rather than geometric mismatch.

\subsubsection{EMA Origin Teacher}
\label{sec:ema_teacher}

Let $f_{\theta}$ be the student and $f_{\xi}$ its EMA teacher, initialized by
$\xi_0=\theta_0$. At iteration $k$, $f_{\theta_{k-1}}$ processes all views,
whereas $f_{\xi_{k-1}}$ processes only \textsc{Origin}. After the optimizer
produces $\theta_k$, the teacher is updated without gradients
\cite{tarvainen2017meanteacher}:
\begin{equation}
    \xi_{k}
    =
    \alpha \xi_{k-1}
    +
    \left(
        1-\alpha
    \right)
    \theta_{k},
    \label{eq:ema_update}
\end{equation}
where $\alpha=0.9998$. The teacher has no optimizer, remains in evaluation
mode, and supplies detached anchors; $\xi_k$ is used at iteration $k+1$.
Temporal averaging stabilizes the \textsc{Origin} reference, while the
\textsc{Origin}-only teacher prevents styled views from shifting the anchor.
Anchoring all styled views to the same \textsc{Origin} creates a star-shaped
alignment structure rather than requiring pairwise agreement among
stylizations. This design scales linearly with the number of sampled views
and avoids treating style-specific generation artifacts as mutual
supervision, keeping consistency tied to the semantically reliable source
observation.

\subsection{Hierarchical Cross-Style Consistency}
\label{sec:hierarchical_consistency}

Let $\mathcal{P}_{\theta}(x)=
\{P_{\theta}^{\ell}(x)\}_{\ell\in\mathcal{L}}$ denote the student FPN
\cite{lin2017fpn}, and define $\mathcal{P}_{\xi}$ analogously. We regularize
these features at scene and object levels.

\subsubsection{Multi-Scale Global Feature Alignment}
\label{sec:global_alignment}

We pool each FPN level, concatenate the results, and normalize:
\begin{equation}
    g_{\theta}(x)
    =
    \operatorname{Norm}
    \left(
        \mathop{\Vert}_{\ell \in \mathcal{L}}
        \operatorname{GAP}
        \left(
            P_{\theta}^{\ell}(x)
        \right)
    \right),
    \label{eq:global_embedding}
\end{equation}
where $\Vert$ is channel-wise concatenation and
$\operatorname{Norm}(z)=z/\lVert z\rVert_2$. With
$d_{\mathrm{cos}}(u,v)=1-u^{\mathsf T}v$, define the detached
\textsc{Origin} anchor
$\bar{g}_i=\operatorname{sg}
\big(g_{\xi}(\widetilde{x}_i^{0})\big)$. For scene set $\mathcal{B}$,
\begin{equation}
    \mathcal{L}_{\mathrm{global}}
    =
    \frac{1}{|\mathcal{B}|m}
    \sum_{i\in\mathcal{B}}
    \sum_{q=1}^{m}
    d_{\mathrm{cos}}
    \big(
        g_{\theta}(\widetilde{x}_i^{s_q}),
        \bar{g}_i
    \big),
    \label{eq:global_loss}
\end{equation}
Here, $\operatorname{sg}$ is identity in the forward pass with zero
derivative. The loss aligns styled student views with the teacher anchor and
excludes the student \textsc{Origin}.

\subsubsection{GT-Guided Instance-Level RoI Alignment}
\label{sec:roi_alignment}

Global pooling may underweight foreground objects. We therefore extract an
embedding at each GT box:
\begin{equation}
    r_{\theta}
    \left(
        x,b_{ij}
    \right)
    =
    \operatorname{Norm}
    \left(
        \operatorname{GAP}
        \left[
            \operatorname{RoIAlign}
            \left(
                \mathcal{P}_{\theta}(x),
                b_{ij}
            \right)
        \right]
    \right).
    \label{eq:roi_embedding}
\end{equation}
Multi-scale RoIAlign \cite{he2017maskrcnn} returns a fixed-resolution FPN
tensor, while shared geometry makes $\widetilde b_{ij}$ identify the same
object across views. Let
$\Omega=\{(i,q,j):i\in\mathcal B,1\le q\le m,
1\le j\le\min(M_i,M_{\max})\}$ and define
$\bar{r}_{ij}=\operatorname{sg}
\big(r_{\xi}(\widetilde{x}_i^{0},\widetilde{b}_{ij})\big)$,
then
\begin{equation}
    \mathcal{L}_{\mathrm{roi}}
    =
    \frac{1}{|\Omega|}
    \sum_{(i,q,j)\in\Omega}
    d_{\mathrm{cos}}
    \big(
        r_{\theta}(\widetilde{x}_i^{s_q},\widetilde{b}_{ij}),
        \bar{r}_{ij}
    \big).
    \label{eq:roi_loss}
\end{equation}
We use $M_{\max}=64$ and match by scene, annotation index, and class,
requiring no proposals, pseudo-labels, logit matching, or box matching.

\subsubsection{Progressive Consistency Regularization}
\label{sec:progressive_consistency}

Early teacher features are unreliable, so we linearly ramp the consistency
weight:
\begin{equation}
    \rho(k)
    =
    \begin{cases}
        0,
        & k \leq k_{\mathrm{s}},\\[3pt]
        \dfrac{k-k_{\mathrm{s}}}
        {k_{\mathrm{e}}-k_{\mathrm{s}}},
        & k_{\mathrm{s}} < k < k_{\mathrm{e}},\\[8pt]
        1,
        & k \geq k_{\mathrm{e}},
    \end{cases},
    \label{eq:consistency_ramp}
\end{equation}
with $k_{\mathrm{s}}=2000$ and $k_{\mathrm{e}}=6000$.

\subsection{Overall Objective and Optimization}
\label{sec:overall_objective}

Every view retains the standard Faster R-CNN loss
$\mathcal L_{\mathrm{det}}=\mathcal L_{\mathrm{roi\text{-}cls}}+
\mathcal L_{\mathrm{roi\text{-}box}}+
\mathcal L_{\mathrm{rpn\text{-}obj}}+
\mathcal L_{\mathrm{rpn\text{-}box}}$. The student objective is
\begin{equation}
    \boxed{
    \mathcal{L}
    =
    \mathcal{L}_{\mathrm{det}}
    +
    \rho(k)
    \left(
        \lambda_{\mathrm{g}}
        \mathcal{L}_{\mathrm{global}}
        +
        \lambda_{\mathrm{r}}
        \mathcal{L}_{\mathrm{roi}}
    \right)
    },
    \label{eq:final_objective}
\end{equation}
with $(\lambda_{\mathrm{g}},\lambda_{\mathrm{r}})=(0.10,0.05)$. After
minimizing Eq.~\eqref{eq:final_objective}, the teacher is updated by
Eq.~\eqref{eq:ema_update}; consistency regularizes rather than replaces
detection supervision. At inference, only the EMA Faster R-CNN is retained;
its detection heads remain unchanged, so OAMSC has the same single-image
computation as the underlying detector.

% \section{Experiments}
% \subsection{DG Benchmark}
% \subsection{Implementation Details}
% \subsection{Results and Comparisons}
% \subsection{Analysis of compatibility}
%写可以兼容不同方法，定性分析
%可以兼容不同检测器，YOLO的展现

% Required packages: amsmath, amssymb, booktabs, graphicx, multirow
% Bibliography file for this section: experiments_references.bib
% Fixed 9-pt Roman table bodies; captions remain controlled by the venue style.
\providecommand{\papertablefont}{\rmfamily\fontsize{9pt}{10.8pt}\selectfont}
\section{Experiments}
\label{sec:experiments}

\subsection{DG Benchmarks}
\label{sec:dg_benchmarks}

We evaluate six source-only DG shifts grouped by their primary domain
change.

\paragraph{Cross-camera shift.}
Cityscapes$\rightarrow$BDD100K daytime-clear evaluates transfer across
datasets and camera systems while controlling time of day and weather
\cite{cordts2016cityscapes,yu2020bdd100k}.

\paragraph{Synthetic-weather shifts.}
Cityscapes$\rightarrow$Foggy Cityscapes and
Cityscapes$\rightarrow$Rainy Cityscapes evaluate robustness to synthesized
fog and rain~\cite{sakaridis2018foggy,hu2019rain}. We remove \emph{train}
from published eight-class Foggy results to match the seven shared classes.

\paragraph{Simulation-to-real shifts.}
Models trained on our 6,000-scene RealDriveSim split
\cite{jadon2025realdrivesim} are evaluated on Cityscapes and BDD100K
daytime-all.

\paragraph{Corruption robustness.}
Cityscapes$\rightarrow$Cityscapes Corruption
\cite{michaelis2019robustness} follows the car-class AP$_{50:95}$ protocol
over 15 corruptions and five severities, where
$\mathrm{rPC}=100\times\mathrm{mPC}/\mathrm{Clean}$. Controlled ablations in
Section~\ref{sec:ablation} instead use seven-class mAP$_{50}$.

\subsection{Implementation Details}
\label{sec:experiment_implementation}

Unless stated otherwise, OAMSC uses Faster R-CNN with an ImageNet-pretrained
ResNet-101-FPN~\cite{he2016resnet} and is trained for 20{,}000 iterations
with a global image batch size of 16. Main results use the 21-view bank and
the EMA model at the fixed 20K endpoint. Cityscapes-source baselines are
published results; RealDriveSim baselines are our reproductions under the
same split, detector, training budget, and evaluation code. Values are
rounded to one decimal, and differences use unrounded results.

\subsection{Results}
\label{sec:experiment_results}

\subsubsection{Cross-Camera and Synthetic-Weather Shifts}

We compare with CDSD~\cite{wu2022cdsd}, SHADE~\cite{zhao2022shade},
SRCD~\cite{rao2024srcd}, FACT~\cite{xu2021fact},
FSDR~\cite{huang2021fsdr}, MAD~\cite{xu2023mad}, and
GDD\textit{(R101)}~\cite{he2025gdd}.

\begin{table*}[!t]
    \centering
    \papertablefont
    \renewcommand{\arraystretch}{1.04}
    \begin{minipage}[t]{0.495\textwidth}
        \centering
        \setlength{\tabcolsep}{2.2pt}
        \begin{tabular}{@{}l c c c c c c c c@{}}
            \toprule
            \textbf{Method} & \textbf{Bike} & \textbf{Bus}
            & \textbf{Car} & \textbf{Mot.} & \textbf{Psn.}
            & \textbf{Rid.} & \textbf{Trk.} & \textbf{mAP} \\
            \midrule
            CDSD  & 22.9 & 20.5 & 33.8 & 14.7 & 18.5 & 23.6 & 18.2 & 21.7 \\
            SHADE & 25.1 & 19.0 & 36.8 & 18.4 & 24.1 & 24.9 & 19.8 & 24.0 \\
            SRCD  & 24.8 & 21.5 & 38.7 & 19.0 & 25.7 & 28.4 & 23.1 & 25.9 \\
            MAD   & --   & --   & --   & --   & --   & --   & --   & 28.0 \\
            GDD\textit{(R101)}   & \textbf{38.4} & 33.4 & 72.0 & \textbf{38.3}
                  & \textbf{60.3} & \textbf{47.0} & \textbf{35.0}
                  & \textbf{46.3} \\
            \midrule
            \textbf{OAMSC}
                  & 34.5 & \textbf{35.3} & \textbf{72.1} & 37.5
                  & 58.0 & 41.2 & 32.0 & 44.4 \\
            \bottomrule
        \end{tabular}
        \caption{Cityscapes$\rightarrow$BDD100K daytime-clear AP$_{50}$
        (\%).}
        \label{tab:city_bdd_results}
    \end{minipage}
    \hfill
    \begin{minipage}[t]{0.495\textwidth}
        \centering
        \setlength{\tabcolsep}{2.2pt}
        \begin{tabular}{@{}l c c c c c c c c@{}}
            \toprule
            \textbf{Method} & \textbf{Bike} & \textbf{Bus}
            & \textbf{Car} & \textbf{Mot.} & \textbf{Psn.}
            & \textbf{Rid.} & \textbf{Trk.} & \textbf{mAP} \\
            \midrule
            FACT  & 31.3 & 27.7 & 35.9 & 23.3 & 26.2 & 41.2 & 13.6 & 28.5 \\
            FSDR  & 34.1 & 36.6 & 43.3 & 27.1 & 31.2 & 44.4 & 19.3 & 33.7 \\
            MAD   & 40.1 & 44.0 & 45.0 & 30.3 & 34.2 & 47.4 & 25.6 & 38.1 \\
            GDD\textit{(R101)}   & 54.2 & 53.8 & 67.5 & \textbf{45.6}
                  & 52.1 & 60.8 & 32.4 & 52.3 \\
            \midrule
            \textbf{OAMSC}
                  & \textbf{56.0} & \textbf{57.4} & \textbf{72.7} & 44.6
                  & \textbf{56.0} & \textbf{61.6} & \textbf{40.7}
                  & \textbf{55.6} \\
            \bottomrule
        \end{tabular}
        \caption{Cityscapes$\rightarrow$Foggy Cityscapes AP$_{50}$ (\%);
        \emph{train} is excluded.}
        \label{tab:city_fog_results}
    \end{minipage}
\end{table*}

\begin{table}[!t]
    \centering
    \papertablefont
    \setlength{\tabcolsep}{3pt}
    \renewcommand{\arraystretch}{1.08}
    \begin{tabular*}{0.62\columnwidth}{
        @{\hspace{6pt}} l @{\extracolsep{\fill}} c @{\hspace{6pt}}
    }
        \toprule
        \textbf{Method} & \textbf{mAP$_{50}$} \\
        \midrule
        FACT & 39.9 \\
        FSDR & 42.8 \\
        MAD & 42.3 \\
        GDD\textit{(R101)} & 57.9 \\
        \midrule
        \textbf{OAMSC (ours)} & \textbf{63.1} \\
        \bottomrule
    \end{tabular*}
    \caption{Cityscapes$\rightarrow$Rainy Cityscapes mAP$_{50}$ (\%);
    classwise AP is unavailable for the compared methods.}
    \label{tab:city_rain_results}
\end{table}

OAMSC reaches $44.4$ mAP$_{50}$ on BDD100K, below GDD's $46.3$ but best on
bus and car (Table~\ref{tab:city_bdd_results}). Under adverse weather, it
surpasses the strongest comparison by $3.2$ points on Foggy Cityscapes and
$5.2$ on Rainy Cityscapes, leading six of seven Foggy categories
(Tables~\ref{tab:city_fog_results} and~\ref{tab:city_rain_results}). The
larger weather gains indicate that appearance diversity is most effective
as visibility departs from the source domain.

\subsubsection{Simulation-to-Real Shifts}

\begin{table*}[!t]
    \centering
    \papertablefont
    \renewcommand{\arraystretch}{1.04}
    \begin{minipage}[t]{0.495\textwidth}
        \centering
        \setlength{\tabcolsep}{2.2pt}
        \begin{tabular}{@{}l c c c c c c c c@{}}
            \toprule
            \textbf{Method} & \textbf{Bike} & \textbf{Bus}
            & \textbf{Car} & \textbf{Mot.} & \textbf{Psn.}
            & \textbf{Rid.} & \textbf{Trk.} & \textbf{mAP} \\
            \midrule
            CDSD  & 36.5 & 41.2 & 54.9 & 31.9 & 41.3 & 53.3 & 18.4 & 39.6 \\
            SHADE & \textbf{40.0} & 46.5 & 54.6 & 30.1
                  & 46.4 & 55.7 & \textbf{22.4} & 42.2 \\
            SRCD  & 38.1 & 46.0 & 52.4 & 28.8 & 43.0 & 53.9 & 22.1 & 40.6 \\
            GDD\textit{(R101)}   & 38.1 & 43.7 & 57.9 & 29.6
                  & \textbf{47.4} & 56.6 & 18.5 & 41.7 \\
            \midrule
            \textbf{OAMSC}
                  & 39.8 & \textbf{50.8} & \textbf{58.7} & \textbf{32.9}
                  & 46.9 & \textbf{56.7} & 18.5 & \textbf{43.5} \\
            \bottomrule
        \end{tabular}
        \caption{RealDriveSim$\rightarrow$Cityscapes AP$_{50}$ (\%) under
        our unified reproduction protocol.}
        \label{tab:realdrivesim_city_results}
    \end{minipage}
    \hfill
    \begin{minipage}[t]{0.495\textwidth}
        \centering
        \setlength{\tabcolsep}{2.2pt}
        \begin{tabular}{@{}l c c c c c c c c@{}}
            \toprule
            \textbf{Method} & \textbf{Bike} & \textbf{Bus}
            & \textbf{Car} & \textbf{Mot.} & \textbf{Psn.}
            & \textbf{Rid.} & \textbf{Trk.} & \textbf{mAP} \\
            \midrule
            CDSD  & 21.9 & 14.8 & 47.5 & 17.9 & 29.1 & 29.1 & 13.3 & 24.8 \\
            SHADE & 19.9 & 18.3 & 47.8 & 18.9 & 29.8 & 29.5 & 15.9 & 25.7 \\
            SRCD  & 17.0 & 9.1  & 45.5 & 15.3 & 22.0 & 27.1 & 12.6 & 21.2 \\
            GDD\textit{(R101)}   & 18.7 & 12.2 & 47.5 & 15.9 & 31.2 & 26.0 & 13.2 & 23.5 \\
            \midrule
            \textbf{OAMSC}
                  & \textbf{23.2} & \textbf{20.9} & \textbf{51.8}
                  & \textbf{23.2} & \textbf{39.3} & \textbf{36.4}
                  & \textbf{21.4} & \textbf{30.9} \\
            \bottomrule
        \end{tabular}
        \caption{RealDriveSim$\rightarrow$BDD100K daytime-all AP$_{50}$
        (\%) under our unified reproduction protocol.}
        \label{tab:realdrivesim_bdd_results}
    \end{minipage}
\end{table*}

OAMSC reaches $43.5/30.9$ mAP$_{50}$ on Cityscapes/BDD100K, improving GDD by
$1.7/7.3$ points and leading four/seven categories, respectively
(Tables~\ref{tab:realdrivesim_city_results}
and~\ref{tab:realdrivesim_bdd_results}). These gains require no architectural
changes to Faster R-CNN.

\subsubsection{Corruption Robustness}

\begin{table*}[!t]
    \centering
    \papertablefont
    \setlength{\tabcolsep}{2pt}
    \begin{tabular}{@{}l c c c c c c c c c c c c c c c c c c@{}}
        \toprule
        \multirow{2}{*}{\textbf{Method}}
        & \multirow{2}{*}{\textbf{Clean}}
        &
        \multicolumn{3}{c}{\textbf{Noise}}
        &
        \multicolumn{4}{c}{\textbf{Blur}}
        &
        \multicolumn{3}{c}{\textbf{Weather}}
        &
        \multicolumn{5}{c}{\textbf{Digital}}
        &
        \multirow{2}{*}{\textbf{mPC}}
        &
        \multirow{2}{*}{\textbf{rPC}}
        \\
        \cmidrule(lr){3-5} \cmidrule(lr){6-9} 
        \cmidrule(lr){10-12} \cmidrule(lr){13-17}
        &
        & Gau.
        & Shot
        & Imp.
        & Def.
        & Gls.
        & Mot.
        & Zm.
        & Snow
        & Frost
        & Fog
        & Bri.
        & Con.
        & Ela.
        & JPG
        & Pix.
        \\
        \midrule
        FSCE~\cite{sun2021fsce}
        & 43.1 & 7.4 & 10.2 & 8.2
        & 23.3 & 20.3 & 21.5 & 4.8
        & 5.6 & 23.6 & 37.1
        & 38.0 & 31.9 & 40.0 & 20.4 & 23.2
        & 21.0 & 48.7
        \\
        OA-Mix~\cite{lee2024oadg}
        & 42.7 & 7.2 & 9.6 & 7.7
        & 22.8 & 18.8 & 21.9 & \textbf{5.4}
        & 5.2 & 23.6 & 37.3
        & 38.7 & 31.9 & \textbf{40.2} & 20.2 & 22.2
        & 20.8 & 48.7
        \\
        OA-DG~\cite{lee2024oadg}
        & \textbf{43.4} & 8.2 & 10.6 & 8.4
        & 24.6 & 20.5 & 22.3 & 4.8
        & \textbf{6.1} & \textbf{25.0} & 38.4
        & 39.7 & 32.8 & \textbf{40.2} & 22.0 & 23.8
        & 21.8 & 50.2
        \\
        GDD\textit{(R101)}~\cite{he2025gdd}
        & 42.1 & 11.0 & 13.6 & \textbf{10.8}
        & 25.0 & 14.2 & 21.4 & 3.4
        & 5.4 & 24.0 & \textbf{39.6}
        & \textbf{40.3} & \textbf{36.3} & 39.2 & 18.9 & 16.0
        & 21.3 & 50.5
        \\
        \midrule
        \textbf{OAMSC}
        & 37.5 & \textbf{12.3} & \textbf{15.0} & 10.0
        & \textbf{25.1} & \textbf{24.6} & \textbf{23.3} & 4.1
        & 6.0 & 20.3 & 34.9
        & 36.5 & 32.4 & 36.0 & \textbf{30.3} & \textbf{33.8}
        & \textbf{23.0} & \textbf{61.3}
        \\
        \bottomrule
    \end{tabular}
    \caption{Car-class AP$_{50:95}$ (\%) on Cityscapes Corruption. Each
    column averages five severities; mPC averages all 15 corruptions.}
    \label{tab:cityscapes_c_results}
\end{table*}

OAMSC achieves $23.0$ mPC and $61.3$ rPC, exceeding GDD by $1.7/10.8$
points (Table~\ref{tab:cityscapes_c_results}). It leads seven corruptions,
notably blur and digital artifacts; GDD is stronger on impulse noise, fog,
brightness, and contrast, while OA methods lead the remainder. Thus, OAMSC
provides complementary robustness with standard detector inference.

\subsection{Compatibility Analysis}
\label{sec:compatibility_analysis}

OAMSC is not tied to a detector paradigm: its multi-style bank and grouped
sampling operate on image--annotation pairs, while feature consistency only
requires intermediate features and GT boxes. We therefore apply the same
training principle to the two-stage Faster R-CNN and the one-stage YOLOv8x
without modifying either inference graph. For this architectural study,
checkpoints are selected on a fixed BDD100K night validation subset.Relative 
to the one-view baselines, 21 views improve YOLOv8x by $2.1$ points
from Cityscapes and $2.5$ from RealDriveSim
(Table~\ref{tab:yolov8x_compatibility}). Together with the Faster R-CNN
results, these gains demonstrate transfer across one- and two-stage
detectors.

\begin{table}[!t]
    \centering
    \papertablefont
    \setlength{\tabcolsep}{2.5pt}
    \renewcommand{\arraystretch}{1.08}
    \begin{tabular*}{0.94\columnwidth}{
        @{\hspace{4pt}} l @{\extracolsep{\fill}} c c c @{\hspace{4pt}}
    }
        \toprule
        \textbf{Source} & \textbf{View count} & \textbf{OAMSC}
        & \textbf{mAP$_{50:95}$} \\
        \midrule
        Cityscapes & 1  & --         & 10.2 \\
        Cityscapes & 13 & \checkmark & 11.9 \\
        Cityscapes & 21 & \checkmark & \textbf{12.3} \\
        \cmidrule(lr){1-4}
        RealDriveSim & 1  & --         & 4.5 \\
        RealDriveSim & 13 & \checkmark & 6.3 \\
        RealDriveSim & 21 & \checkmark & \textbf{7.0} \\
        \bottomrule
    \end{tabular*}
    \caption{Transfer to the one-stage YOLOv8x detector on BDD100K
    night-all. View count includes \textsc{Origin}; the one-view setting is
    the source-specific baseline.}
    \label{tab:yolov8x_compatibility}
\end{table}

OAMSC primarily improves appearance robustness through label-preserving
source expansion and auxiliary feature regularization, without replacing
the detector architecture or its detection objective. This modular design
permits integration with feature-, optimization-, and architecture-based DG
methods in principle, offering a possible route to broader robustness across
shifts. The effectiveness of such combinations requires empirical
validation.

%--------------------------------------------------
%下面是第5章的图片，放在这里不会超页
\begin{figure*}[!ht]
    \centering
    \includegraphics[
        width=0.95\textwidth,
        keepaspectratio
    ]{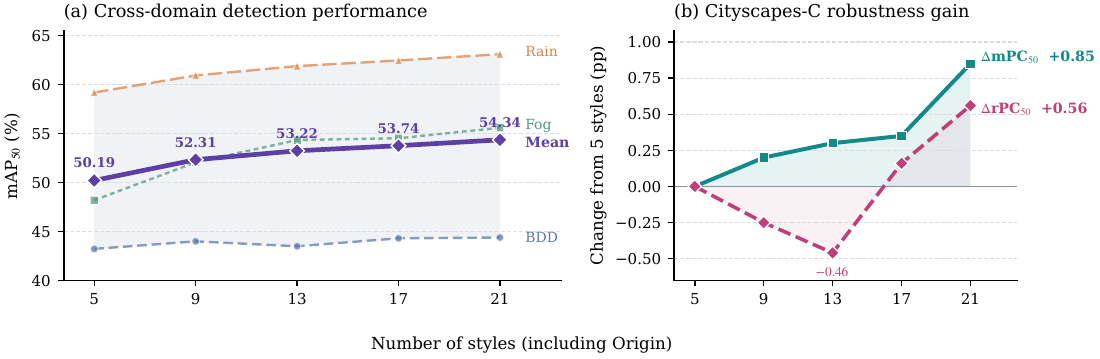}
    \caption{Effect of style quantity (including \textsc{Origin}).
    (a) Target-domain mAP$_{50}$ and its mean (bold); shading spans the three
    domains. (b) Cityscapes-C gains over the 5-style setting.}
    \label{fig:ablation_style_quantity}
\end{figure*}
%--------------------------------------------------

%-----------------------------------------------------------
% Required packages: amsmath, amssymb, booktabs, graphicx
% Fixed 9-pt Roman table bodies; captions remain controlled by the venue style.
\providecommand{\papertablefont}{\rmfamily\fontsize{9pt}{10.8pt}\selectfont}
\section{Ablation Studies}
\label{sec:ablation}

We ablate appearance diversity, hierarchical consistency, and
\textsc{Origin}-anchored training under a common protocol: Faster R-CNN with
ResNet-101-FPN, Cityscapes and its corresponding style variants as the
training data, identical optimization, and the 20{,}000-iteration
checkpoint. Evaluation covers BDD100K daytime clear (BDD), Foggy Cityscapes
(Fog), Rainy Cityscapes (Rain), and Cityscapes-C. In the component tables,
mAP$_{50}$ is the mean over BDD, Fog, and Rain. Cityscapes-C mPC$_{50}$
averages 15 corruptions at five severities, and rPC is
$100\times\mathrm{mPC}_{50}/\mathrm{mAP}_{50}^{\mathrm{clean}}$. All values
are percentages; the 17-style point, L11, and M3 denote the same model under
different ablation views.

\subsection{Style Quantity}
\label{sec:ablation_style_quantity}

We use nested banks of 5--21 views per scene, counting \textsc{Origin}, while
always sampling one \textsc{Origin} and three styles per group. This varies
appearance coverage without changing per-iteration cost.

From 5 to 21 styles, mean mAP$_{50}$ rises by $4.15$ points; gains are largest
on Fog ($+7.39$) and Rain ($+3.91$), versus $+1.16$ on BDD
(Fig.~\ref{fig:ablation_style_quantity}). Cityscapes-C mPC$_{50}$ increases
monotonically by $0.85$ points. rPC first drops to $-0.46$ at 13 styles, then
recovers to $+0.56$. Thus, broad coverage mainly improves adverse-domain and
corruption robustness, with 21 styles giving the strongest aggregate result.
Within the evaluated 5--21-style range, both the target-domain mean and
Cityscapes-C mPC continue to rise without evident saturation. This supports
the inference that broader, label-preserving appearance coverage produces
progressively more robust and transferable representations in the studied
range, although it does not establish that gains will persist beyond 21
styles.

\subsection{Hierarchical Consistency}
\label{sec:ablation_consistency_losses}

With 17 styles, we independently toggle global FPN consistency (G) and
GT-RoI consistency (R), retaining grouped sampling, synchronized geometry,
and the EMA teacher.

\begin{table}[!ht]
    \centering
    \papertablefont
    \setlength{\tabcolsep}{2pt}
    \renewcommand{\arraystretch}{1.08}
    \begin{tabular*}{0.92\columnwidth}{
        @{\hspace{4pt}} c @{\extracolsep{\fill}} c c c c @{\hspace{4pt}}
    }
            \toprule
            \textbf{Setting}
            & \textbf{G}
            & \textbf{R}
            & \textbf{mAP$_{50}$}
            & \textbf{\shortstack{Cityscapes-C\\rPC}}
            \\
            \midrule
            L00 & -- & --
            & 53.56 & 63.23
            \\
            L10 & \checkmark & --
            & 53.61 & 63.48
            \\
            L01 & -- & \checkmark
            & 53.56 & 63.23
            \\
            L11 & \checkmark & \checkmark
            & \textbf{53.74} & \textbf{63.51}
            \\
            \bottomrule
    \end{tabular*}
    \caption{Global (G) and GT-RoI (R) consistency with 17 styles.
    Best results are bold.}
    \label{tab:ablation_consistency_losses}
\end{table}

G alone improves mAP$_{50}$/rPC by $0.05/0.25$ points, whereas R alone is
neutral after rounding. Their combination is best, improving L00 by
$0.18/0.28$ points (Table~\ref{tab:ablation_consistency_losses}). The losses
are therefore complementary refinements, not the primary source of gain.

\subsection{Grouped Sampling and EMA Anchoring}
\label{sec:ablation_grouping_ema}

M0 uses random sampling and detection loss; M1 adds scene grouping; M2 uses
the current student \textsc{Origin} as an online anchor; M3 replaces it with
the detached EMA anchor.

\begin{table}[!ht]
    \centering
    \papertablefont
    \setlength{\tabcolsep}{2pt}
    \renewcommand{\arraystretch}{1.08}
    \begin{tabular*}{0.98\columnwidth}{
        @{\hspace{4pt}} c @{\extracolsep{\fill}} l c c @{\hspace{4pt}}
    }
            \toprule
            \textbf{Setting}
            & \textbf{Configuration}
            & \textbf{mAP$_{50}$}
            & \textbf{\shortstack{Cityscapes-C rPC}}
            \\
            \midrule
            M0 & Random; det.\ only
            & 50.02 & 59.36
            \\
            M1 & Grouped; det.\ only
            & 52.50 & 60.25
            \\
            M2 & Grouped; online
            & 52.35 & 60.22
            \\
            M3 & Grouped; EMA
            & \textbf{53.74} & \textbf{63.51}
            \\
            \bottomrule
    \end{tabular*}
    \caption{Scene grouping and \textsc{Origin} anchoring with 17 styles.
    Best results are bold.}
    \label{tab:ablation_grouping_ema}
\end{table}

Grouping alone yields $+2.48$ mAP$_{50}$ and $+0.89$ rPC points over M0
(Table~\ref{tab:ablation_grouping_ema}). The online anchor slightly degrades
both metrics, while EMA improves M2 by $1.39/3.29$ points and M0 by
$3.72/4.15$ points. Exact correspondence is thus valuable, but a temporally
stable anchor is necessary to exploit it reliably.

\subsection{Discussion and Limitations}
\label{sec:ablation_discussion}

The ablations identify appearance coverage and scene grouping as the main
contributors, with EMA providing a stable anchor and global/RoI consistency
offering smaller complementary gains. OAMSC is therefore a coupled
data-organization and asymmetric-learning strategy rather than an isolated
feature penalty.

Offline multi-style banks increase generation, storage, I/O, and training
costs with the number of source scenes and retained styles. OAMSC also
requires reliable cross-view label consistency, such that the original
annotations remain valid for every generated view. Inference remains
unchanged, and this training-only overhead may be acceptable in
safety-critical applications when justified by robustness gains. Current
evidence remains limited in architectures and repeated runs; broader
validation should include multiple seeds and detectors, as well as adaptive
weighting and reliability-aware anchors for noisy sources.

%------------------------------------------------------
\section{Conclusion}
\label{sec:reference_examples}

We presented Origin-Anchored Multi-Style Consistency (OAMSC), a source-only
DG framework that learns from label-consistent appearance variants while
retaining standard detector inference. Experiments across cross-dataset,
adverse-weather, simulation-to-real, and corruption shifts demonstrate its
robustness, particularly under substantial appearance changes.

The ablations further show that appearance coverage and same-scene grouping
provide the principal gains, EMA stabilizes the semantic reference, and
hierarchical consistency offers complementary refinement. These findings
suggest that robust generalization depends on jointly structuring appearance
variation, cross-view correspondence, and reference stability rather than
adding an isolated regularization term. Future work will develop stronger
consistency objectives and adaptive alignment strategies to better exploit 
same-scene correspondence across appearance domains and further improve transfer 
to unseen domains.

\clearpage
\bibliography{aaai2027}


\end{document}